\documentclass[letterpaper, 10 pt, conference]{ieeeconf}  
\IEEEoverridecommandlockouts % This command is only needed if                                       % you want to use the \thanks command

\usepackage{graphicx} 
\usepackage{caption}
\usepackage[caption=false,font=footnotesize]{subfig}
\usepackage{times} 
\usepackage{amsmath}
\usepackage{amssymb} 
\usepackage[english]{babel}
\usepackage{algorithm2e}
\usepackage{balance}
\usepackage{xcolor}
\usepackage{booktabs}
\usepackage{hyperref}
\usepackage{cite}
\usepackage{multirow}
\usepackage{colortbl}

\title{\LARGE \bf
\vspace{-1.5em} % Subimos un poco para aprovechar margen
{\normalfont \scriptsize \centering
Published in \textit{IEEE Robotics and Automation Letters}, vol. 11, no. 1, pp. 169--176 (2026). \\
DOI: \href{https://doi.org/10.1109/LRA.2025.3632615}{10.1109/LRA.2025.3632615}
\par \vspace{1em}} % Espacio antes del título real
D-LIO: 6DoF Direct LiDAR-Inertial Odometry based on Simultaneous Truncated Distance Field Mapping
}

\author{L. Coto-Elena$^{1}$, J.E. Maese$^{1}$,  L. Merino$^{1}$ and F. Caballero$^{1}$% <-this % stops a space
\thanks{*This work is partially supported by the grants INSERTION (PID2021-127648OB-C31) and NORDIC (TED2021-132476B-I00), both funded by the ``Agencia Estatal de Investigación -- Ministerio de Ciencia, Innovación y Universidades'' and the ``European Union NextGenerationEU/PRTR''.}% <-this % stops a space
\thanks{$^{1}$The authors are with the Service Robotics Laboratory, Universidad Pablo de Olavide, Seville, Spain. {\tt\small \{lcotele,jemaealv,lmercab,fcaballero\}@upo.es}}%
}

\newcommand{\rev}[1]{\textcolor{black}{#1}}

\begin{document}

\maketitle
\thispagestyle{empty}
\pagestyle{empty}

%%%%%%%%%%%%%%%%%%%%%%%%%%%%%%%%%%%%%%%%%%%%%%%%%%%%%%%%%%%%%%%%%%%%%%%%%%%%%%%%
\begin{abstract}
This paper presents a new approach for 6DoF Direct LiDAR-Inertial Odometry (D-LIO) based on the simultaneous mapping of truncated distance fields on CPU. Such continuous representation (in the vicinity of the points) enables working with raw 3D LiDAR data online, avoiding the need of LiDAR feature selection and tracking, simplifying the odometry pipeline and easily generalizing to many scenarios. The method is based on the proposed Fast Truncated Distance Field (Fast-TDF) method as a convenient tool to represent the environment, employing binary masks that encodes the L1 distance. Such representation enables i) solving the LiDAR point-cloud registration as a nonlinear optimization process without the need of selecting/tracking LiDAR features in the input data, ii) simultaneously producing an accurate truncated distance field map of the environment, and iii) updating such map at constant time independently of its size. The approach is tested using open datasets, aerial and ground. It is also benchmarked against other state-of-the-art odometry approaches, demonstrating the same or better level of accuracy with the added value of an online-generated TDF representation of the environment, that can be used for other robotics tasks as planning or collision avoidance. The source code is publicly available at \url{https://github.com/robotics-upo/D-LIO.git}

\end{abstract}

%%%%%%%%%%%%%%%%%%%%%%%%%%%%%%%%%%%%%%%%%%%%%%%%%%%%%%%%%%%%%%%%%%%%%%%%%%%%%%%%
\section{Introduction}
\label{sec:intro}

Accurate vehicle localization is a crucial aspect of robotics, directly influencing autonomous navigation, remote exploration, and other advanced applications. Various techniques are employed to improve localization, combining data from different sensors such as cameras, inertial measurement units (IMUs), LiDAR and radar \cite{liosam2020shan}. LiDAR-based approaches \cite{KISSICP} are widely used due to their capacity to provide high-density and accurate 3D data unaffected by lighting conditions. These techniques have been proven to provide accurate positioning while also enabling environmental mapping \cite{LOAM}\cite{9681177}. However, they have certain limitations. In feature-sparse environments, traditional feature-based techniques \cite{LeGOLOAM}\cite{9636655} may struggle to extract reliable information, leading to suboptimal localization. Additionally, when dealing with extremely dense point clouds, the computational cost increases significantly, making real-time execution challenging.

Beyond the choice of sensors, the underlying map representation plays a pivotal role in the accuracy and efficiency of localization systems. Most of the state-of-the-art LiDAR SLAM approaches represent maps in the form of point clouds (3D points and/or features) together with ICP-like methods for LiDAR registration \cite{GICP}\cite{KISSICP}. While effective, these discrete representations often face challenges in scalability, sparsity, and differentiability \cite{Voxfield}\cite{OpenVDB}. To address these limitations, distance fields, particularly Signed and Truncated Distance Fields \cite{Curless1996A}, have emerged as a promising alternative. %powerful alternative to traditional occupancy grids or feature maps, offering a continuous and differentiable encoding of spatial structure. 
These volumetric representations capture obstacle proximity in a continuous fashion, which is especially beneficial for point cloud registration tasks by providing smooth gradients and implicit surface descriptions. However, the application to LiDAR SLAM is still limited to relatively small volumes (rooms, house floor, etc) or with large quantization errors that prevent their use in general purpose application \cite{iSDF}\cite{ HIO-SDF}. As a result, most LiDAR Distance Field approaches in the state of the art focus on mapping, assuming perfect sensor localization \cite{VDB-GPDF}\cite{Voxfield}. 

This paper presents, to the best of our knowledge, the first 6DoF LiDAR-Inertial odometry method based on simultaneous mapping of truncated distance field in large volumes. Although the approach has potential for parallelization, it is able to map the environment and estimate the odometry approaching real time using just CPU. Unlike learning-based methods, the proposed Fast-TDF approach provides an accurate representation of the environment from the first scan.

% Among them, Fast Truncated Distance Fields \cite{KinectFusion} offer a promising trade-off between computational efficiency and representational power. By constraining the computation to a fixed influence region around each surface and employing binary masks to update only those voxels within this band, Fast-TDFs significantly reduce memory and processing overhead, enabling real-time updates and fast access operations. This makes them particularly suitable for online localization and odometry tasks, where timely responses and scalability are essential.

% Revisar 
%In this paper, a direct odometry approach based on fast-truncated distance fields is proposed to efficiently process LiDAR point clouds, overcoming the limitations of traditional methods to improve computational efficiency and trajectory accuracy, enhancing autonomous navigation across diverse environments. 

%The proposed algorithm has shown improvements over existing methods such as KISS-ICP, M-LOAM, and FAST-LIO2 when evaluated on the VIRAL and College datasets, consistently outperforming them, while obtaining comparable results in some specific trajectories to A-LOAM.

The paper is structured as follows. First, Section \ref{sec:soa} analyzes the current state of the art on LiDAR odometry and distance field computation. Later, the proposed Fast Truncated Distance Field method is presented in Section \ref{sec:ftedf}. Then, Section \ref{sec:dlo} details the LiDAR-Inertial odometry approach based on simultaneous TDF mapping. The proposed methods are tested and benchmarked using public datasets and open-sourced LiDAR odometry approaches in Section \ref{sec:exp}. Finally, Section \ref{sec:conclusions} presents the conclusions and future work.

%%%%%%%%%%%%%%%%%%%%%%%%%%%%%%%%%%%%%%%%%%%%%%%%%%%%%%%%%%%%%%%%%%%%%%%%%%%%%%%%%%%%%%%%%%%%%%%%%%%%%%%%%%%%%%%%%%
 \section{State of the Art}
\label{sec:soa}
\subsection{LiDAR-based Odometry}

Modern LiDAR-based odometry relies on high-resolution 3D point clouds to estimate motion, offering a key advantage over cameras by operating independently of ambient lighting conditions. 
%Traditional methods, such as ICP \cite{121791} and NDT \cite{1249285} register consecutive scans by minimizing alignment errors. These have been enhanced by approaches like GICP \cite{GICP}, % TrICP \cite{1047997},CT-ICP \cite{9811849} and KISS-ICP \cite{KISSICP} that improve robustness and computational efficiency. Additionally, scan-to-map strategies, including IMLS-SLAM \cite{8460653} and DLO \cite{9681177}, leverage accumulated map information to enhance localization stability.
Traditional methods, such as ICP \cite{121791} and NDT \cite{1249285}, register consecutive scans by minimizing alignment errors. \rev{These have been enhanced by variants like GICP \cite{GICP}, which boost robustness and efficiency. Advanced methods such as CT-ICP \cite{9811849}, KISS-ICP \cite{KISSICP}, DLO \cite{9681177} and DLIO \cite{chen2023directlidarinertialodometrylightweight} all perform scan-to-local-map (or scan-to-submap) registration, leveraging accumulated map information for improved stability.} To reduce computational cost, feature-based methods extract geometric structures rather than using entire point clouds. 
LOAM \cite{LOAM} pioneered this approach by detecting edges and planar features, later extended by LeGO-LOAM\cite{LeGOLOAM}, \rev{F-LOAM \cite{9636655}, and A-LOAM \cite{9963800}, while M-LOAM \cite{M-LOAM} adapts the framework to multi-LiDAR configurations. 
In contrast, SuMa++ \cite{SuMa} abandons sparse features in favor of a dense surfel map, each surfel encoding position, normal, temporal stability, and semantic information, for projective ICP registration and dynamic-object filtering.}

Deep learning-based odometry has emerged to address challenges in feature consistency and matching. LO-Net \cite{LO-Net} improves scan alignment by predicting surface normals and enforcing geometric constraints, while LodoNet \cite{Zheng_2020} uses keypoint-based pose estimation. Unsupervised learning techniques, such as VertexNet \cite{9197366}, model pose uncertainty to enhance accuracy across diverse environments. These methods aim to improve generalization and robustness, complementing traditional LiDAR techniques.

Despite its advantages, LiDAR-only odometry faces several limitations. In featureless environments, such as tunnels, point cloud registration becomes unreliable due to insufficient constraints. Additionally, LiDAR’s low frame rate makes it less effective for capturing rapid motion dynamics. To mitigate these issues, sensor fusion with complementary modalities is employed to enhance robustness and computational efficiency. LiDAR-Inertial odometry is an approach commonly used as IMU provide high-frequency inertial measurements capturing short-term motion dynamics with minimal latency. 

Sensor fusion methods can be categorized depending on how each sensor contributes to the estimation of the pose.  In loosely coupled approaches, LiDAR and IMU measurements are processed independently, and their outputs are fused at a later stage. These methods offer modularity and robustness, but may not fully exploit the complementary nature of both sensors. Examples include LOAM \cite{LOAM} or Lego-LOAM \cite{LeGOLOAM}. 

On the other hand, tightly coupled approaches directly integrate raw IMU and LiDAR measurements within the same optimization framework, enabling continuous state estimation and higher accuracy. \rev{Methods such as LIO-SAM \cite{liosam2020shan}, FastLIO2 \cite{FastLIO2} and Faster-LIO \cite{Faster-LIO}  perform well in feature-sparse environments, where IMU constraints help maintain reliable pose estimates. }Despite advances in LiDAR-based odometry, several challenges still hinder its performance. One of the main issues is the high computational cost of processing dense point clouds, making it difficult to execute registration and optimization algorithms efficiently on hardware-constrained platforms. Another significant challenge is the difficulty of extracting discriminative features in homogeneous environments or repetitive geometries, potentially leading to errors in motion estimation. In such scenarios, direct odometry methods, that leverage the full point cloud information without relying on feature extraction, are often considered as an alternative \cite{9681177} \cite{SuMa}. Furthermore, filtering and data reduction strategies must balance the need to decrease computational load while preserving essential information for localization.  
\rev{Recently, neural implicit representations have been explored as an alternative to traditional geometric or TDF-based maps. 
Approaches such as NeRF-LOAM \cite{deng2023nerfloamneuralimplicitrepresentation}, PIN-SLAM \cite{Pan_2024}, and KN-LIO \cite{wang2025knliogeometrickinematicsneural} jointly optimize odometry and dense mapping within continuous neural fields. 
These methods leverage signed distance functions or neural point-based representations to achieve accurate pose estimation together with high-fidelity 3D reconstructions, often integrating LiDAR and IMU measurements. 
Although promising, they typically involve high computational demands and GPU acceleration, which may limit their applicability in resource-constrained robotic platforms.}

\subsection{Distance Fields Computation}

% Volumetric mapping and Simultaneous Localization and Mapping (SLAM) methods have significantly contributed to robot navigation, planning, and localization across diverse environments. These approaches primarily utilize discrete voxel-based structures, hierarchical data representations, probabilistic models, and implicit neural representations.

Volumetric mapping and SLAM methods primarily utilize discrete voxel grids, hierarchical data structures, probabilistic models, and more recently implicit neural fields.

Voxel-based methods are extensively used due to their simplicity and ease of implementation. Voxblox \cite{Voxblox} efficiently generates Euclidean Signed Distance Fields (ESDF) from Truncated Signed Distance Fields (TSDF), establishing a common baseline in volumetric mapping. % FIESTA \cite{FIESTA} further enhances computational efficiency by employing occupancy-based wavefront propagation instead of TSDF, thereby reducing complexity. 
Additionally, Voxfield \cite{Voxfield} refines accuracy through non-projective TSDF fusion complemented by normal estimation from input point-cloud data. \rev{Notably, both Voxblox and Voxfield use voxel hashing, which enables sparse spatial memory allocation and helps mitigate scalability issues in large-scale environments. However, these approaches can still suffer from discretization artifacts and access-time variability, particularly in CPU-only systems or high-frequency mapping tasks.}
% Despite their popularity and CPU efficiency, these voxel-based solutions exhibit inherent limitations such as scalability issues and discretization errors, reducing their effectiveness and precision in real-time SLAM applications.

%Hierarchical data structures such as OpenVDB \cite{OpenVDB} have been proposed to mitigate scalability constraints. Approaches leveraging OpenVDB include VDB-EDT \cite{VDB-EDT}, which represents ESDF through efficient distance transforms; \rev{or VDBFusion \cite{VDBFusion}, which integrates TSDF data into OpenVDB but does not explicitly produce ESDFs.} %or VDBblox \cite{VDBblox}, which enhances the propagation efficiency of ESDF within OpenVDB. 
%While these hierarchical methods improve scalability and memory usage, their discrete nature still limits smooth interpolation capabilities and continuous optimization.

Hierarchical data structures such as OpenVDB \cite{OpenVDB} have been proposed to mitigate scalability constraints. Approaches leveraging OpenVDB include \rev{VDBFusion \cite{VDBFusion}}, which integrates TSDF data into OpenVDB. While these hierarchical methods improve scalability and memory usage, their discrete nature still limits smooth interpolation capabilities and continuous optimization.

%Implicit neural representations have emerged as continuous alternatives to discrete methods, providing dense and differentiable volumetric maps. DeepSDF \cite{DeepSDF} introduced a neural-network-based implicit model to represent continuous SDFs. Continual Neural Mapping \cite{Continual_neural_mapping} proposed an incremental neural method employing replay buffers, but real-time incremental updates were first effectively demonstrated by iSDF \cite{iSDF}. More recently, HIO-SDF \cite{HIO-SDF} introduced a hierarchical approach combining global voxel structures with local LiDAR data to train neural models online. \rev{SHINE-Mapping \cite{SHINE} employs a sparse-octree neural SDF for large-scale reconstruction, but it targets mapping rather than online LiDAR–inertial odometry.} TNDF-Fusion \cite{TNDFFusion} compresses city-scale LiDAR maps via a Tri-Pyramid TNDF and NeRF-SLAM \cite{NerfSlam} delivers real-time dense monocular SLAM with hierarchical NeRFs, yet none of them unifies full 6-DoF LiDAR–inertial odometry with on-the-fly distance field mapping on a CPU. Despite these advances, neural methods typically require significant pre-training, heavy computation, and are limited to small-scale SLAM due to memory and inference constraints.

Implicit neural representations have emerged as continuous and differentiable alternatives to discrete volumetric maps. DeepSDF \cite{DeepSDF} introduced neural SDFs for continuous surface modeling, and real-time incremental updates were first achieved by iSDF \cite{iSDF}. More recently, HIO-SDF \cite{HIO-SDF} combines global voxel structures with local LiDAR data for online training, while \rev{SHINE-Mapping \cite{SHINE} employs a sparse-octree neural SDF for large-scale reconstruction, but it targets mapping rather than online LiDAR–inertial odometry.} NeRF-SLAM \cite{NerfSlam} delivers real-time dense monocular SLAM with hierarchical NeRFs, yet none of them unifies full 6-DoF LiDAR–inertial odometry with on-the-fly distance field mapping on a CPU. Despite these advances, neural methods typically require significant pre-training, heavy computation, and are limited to small-scale SLAM due to memory and inference constraints.

% Gaussian Processes (GP) have been explored as probabilistic alternatives for representing distance fields, offering uncertainty estimation capabilities. Previous approaches have used GP-based models for uncertainty-aware mapping \cite{GPDF_for_mapping}, but their inference remains computationally intensive due to the inversion of covariance matrices. To address computational constraints, hierarchical representations such as Octrees have been employed \cite{GPDF_for_mapping}. VDB-GPDF \cite{VDB-GPDF} leverages a logarithmic Gaussian Processes embedded within the OpenVDB structure, generating continuous ESDFs with uncertainty measures. Nonetheless, the computational complexity associated with GP inference remains a significant obstacle for real-time implementation.

Gaussian Processes (GP) have been proposed as probabilistic distance‐field representations with built-in uncertainty estimation. GP-based mapping methods remain computationally expensive due to covariance matrix inversion, even when hierarchical structures such as Octrees are used. VDB-GPDF \cite{VDB-GPDF} embeds logarithmic GP models in OpenVDB to produce continuous ESDFs with uncertainty, but GP inference still limits real-time performance.

A significant limitation among these methodologies is their primary orientation toward mapping tasks, with limited or no direct application to real-time SLAM due to computational demands and quantization errors inherent in discretized representations. Consequently, real-time trajectory estimation and incremental map updates remain challenging.

In contrast, the approach proposed in this paper, Direct LiDAR-Inertial Odometry (D-LIO), explicitly targets these limitations. D-LIO simplifies the odometry pipeline by eliminating the necessity for explicit LiDAR feature selection and tracking. Moreover, it maintains the environment map update process in constant time, irrespective of map size, significantly enhancing scalability. By directly fusing raw LiDAR point-cloud data into an optimized Fast Truncated Distance Field (Fast-TDF) representation, our method eliminates the additional computational overhead typically required to convert TSDF to ESDF, thus facilitating real-time integration into SLAM systems.

%%%%%%%%%%%%%%%%%%%%%%%%%%%%%%%%%%%%%%%%%%%%%%%%%%%%%%%%%%%%%%%%%%%%%%%%%%%%%%%%
\section{Fast Truncated Distance Field}
\label{sec:ftedf}

%Distance Fields (DF) have been successfully used as an efficient environment representation for both robot navigation and localization approaches [REF][REF]. In general, the EDF is represented as a grid map (2D or 3D) that encodes the distance to the closest obstacle in the environment in each grid cell. Thus, if we need to compute the distance to the closest obstacle at a given position in the environment, we only have to consult the distance value stored in the grid map in such position.

%However, the computational effort required to estimate the DF in large environments is very high, preventing its use online. Some approaches exploit the particularities of the problem to solve to achieve online computation, as in re-planning, where the region to compute DF use to be small [REF]. But this kind of advantages cannot be considered in localization/odometry approaches, where the whole environment is subject to be explored by the robot. Another important drawback of EDFs is their global nature. Adding just a single cell with an obstacle might need to update a large portion of the EDF because the area of influence of every obstacle cell depends on the location of the other obstacles. 

%Truncated Distance Fields (TDF) are a good choice in terms of computation because we do not need to evaluate all the space in the environment, only the cells under a given distance from obstacles. This also benefits grid cell updating with new obstacles, because new obstacles affect only environment cells under the truncation distance. 

Distance Fields (DFs) are able to provide the \rev{distance to the} closest obstacle \rev{from any} point in space, no matter how far we are from the obstacle. On the other hand, Truncated Distance Fields (TDFs) are only useful when the coordinate has an obstacle under its truncation distance. While this might be seen as a great advantage of DFs for localization, practical aspects limits and even dismiss such advantage because far points are normally rejected by the outlier rejection process during point cloud registration. Thus, if the truncation distance is properly tuned according to the localization approach at hand, TDFs might be a very efficient representation of the distance field.

This paper proposes using TDF to estimate the distance field. The proposed Fast-TDF takes advantage of binary masks to compute the $L_1$ distance into the area of influence of each obstacle. We will see how using binary masks to encode TDFs is very convenient, changing the computational paradigm from map size dependence to point cloud size dependence, so that the computation will depend on the number of points to be included into the map, not the map size.

\subsection{Binary kernel}
\label{binarykernel}
\begin{figure}[t]
    \centering
    \subfloat[]{\includegraphics[width=0.29\columnwidth]{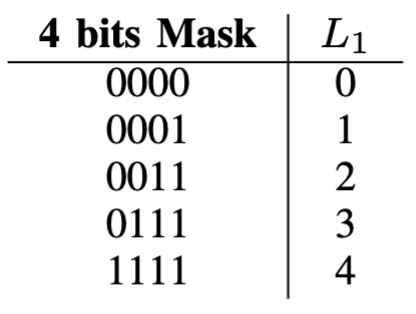}}
    \hspace{1cm}
    \subfloat[]{\includegraphics[width=0.29\columnwidth]{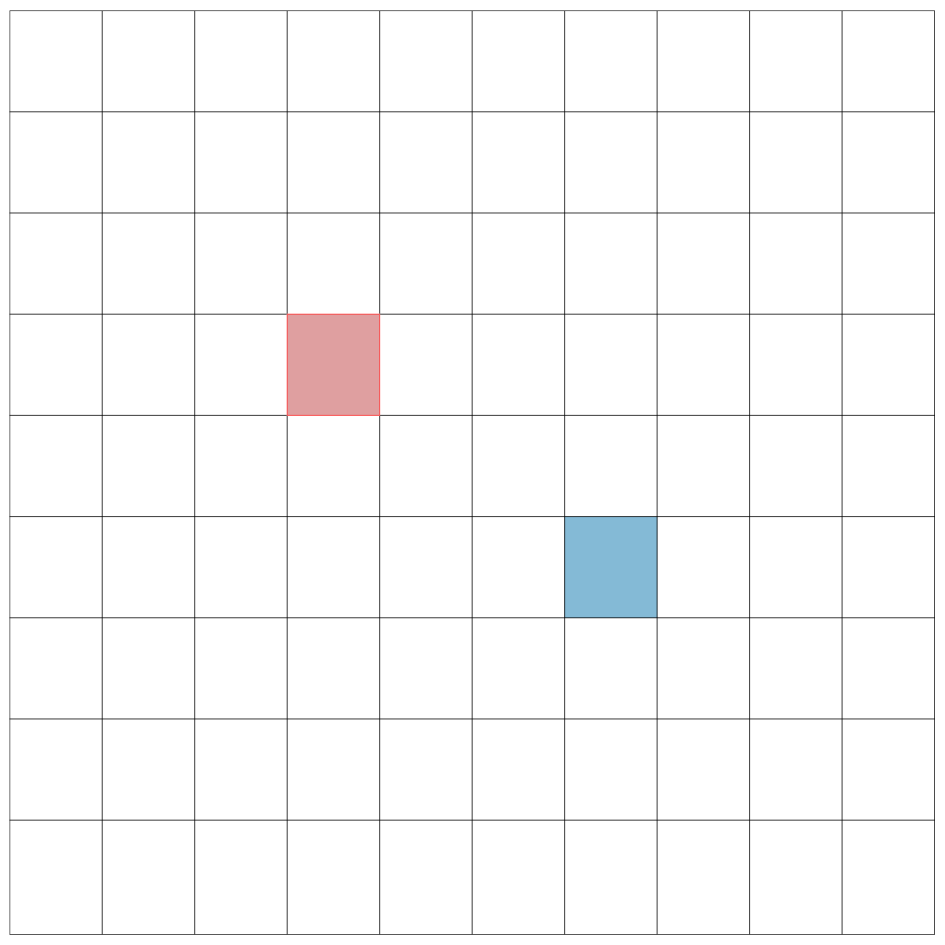}}\\
    \subfloat[]{\includegraphics[width=0.4\columnwidth]{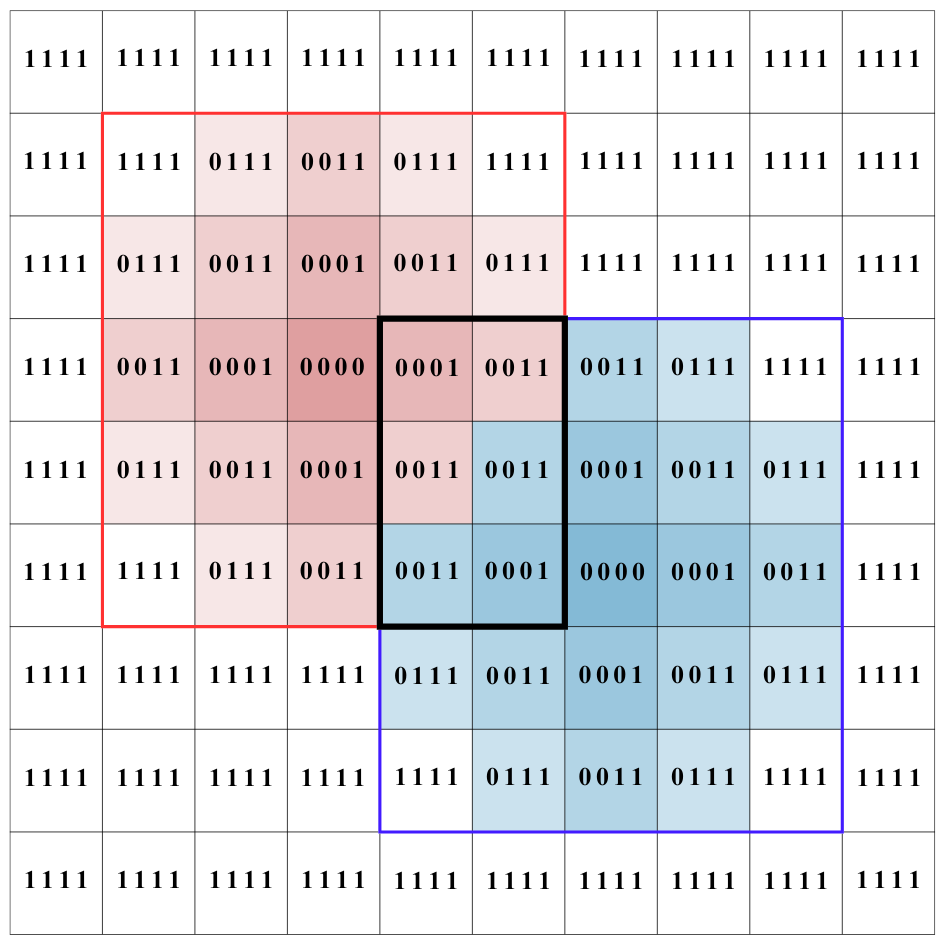}}
    \caption{2D example of $5 \times 5$ binary kernel using 4 bits encoding. (a) Correspondence between mask and $L_1$ distance. (b) 2D occupancy representation of the input point cloud over the grid. (c) Merged distance map resulting from the bitwise-AND operation over overlapping binary kernels centered at each occupied cell.}
    \label{fig:kernel_example}
\end{figure}

Our Fast-TDF makes use of a binary kernel to compute the distance. This kernel must be applied to the map in each occupied cell. As a result, it will produce the distance of the surrounding free cells to the closest occupied under the truncation distance. The kernel represents the $L_1$ distance from each cell to an occupied cell in the center of the kernel. The distances into the kernel are expressed as bit-set masks following these conventions (Fig. \ref{fig:kernel_example}.(a) shows an example of four bits):
\begin{itemize}
    \item A zeroed mask indicates $L_1$ distance equal to zero.
    \item Starting with the Least Significant Bit (LSB), each bit set accounts for the distance between two consecutive cells.
    \item An all-one mask indicates the maximum $L_1$ distance, which is equal to the number of bits in the mask.
\end{itemize}

This distance encoding has a great advantage, we can compute the shortest distance into an arbitrary set by means of the bitwise-AND operation. No matter the number of distances or the values involved, a bitwise-AND among all the distances will provide the shortest value. This is positive at three levels: i) The bitwise-AND operation can be computed extremely fast in computers, with both, scalar and vectorized instructions, ii) We do not need to perform value comparisons at all, with the corresponding computation impact, and iii) The distance insertion order does not affect the result, because the bitwise-AND operation is commutative, which simplifies the parallelism.

As an illustrative example, let's assume we have three different distance values represented as a bit-set mask of 8 bits, these values are $00111111$ ($L_1=6$), $00001111$ ($L_1=4$) and $00000001$ ($L_1=1$). The shortest distance among these values can be easily estimated by computing the bitwise-AND operation without performing value comparison: 
\begin{equation*}
00111111\quad\&\quad00001111\quad\&\quad00000001 = 00000001
\end{equation*}

%Figure \ref{fig:kernel_example} shows a 2D example of a $5 \times 5$ binary kernel using four bits encoding. We assume the center of the kernel is in coordinates $(0,0)$, so kernel corners are in $(2, 2)$, $(-2, 2)$, $(2, -2)$ and $(-2, -2)$, which gives us $L_1 = 4$ in the corners. It can be seen that the center has mask value $0000$, that is $L_1=0$, while the corners have mask value $1111$ representing $L_1=4$.

Figure \ref{fig:kernel_example} illustrates a 2D example where two binary kernels of size  $5 \times 5$ with 4 bit masks encoding are centered, respectively, on a red and a blue cell (Fig. \ref{fig:kernel_example}.(b)). The bit mask encodes the $L_1$ distance, from $0000$ at the kernel center ($L_1 = 0$) to $1111$ at the corners ($L_1 = 4$), as shown in Fig. \ref{fig:kernel_example}.(a). In Fig. \ref{fig:kernel_example}.(c), both kernels are applied using bitwise-AND around their respective centers. In the overlapping region (outlined in black), we can see how the method preserves the shortest distance to the nearest center. All map cells were initialized to $1111$ (the truncation distance) before applying the binary kernels, indicating the maximum or unknown distance. 

\subsection{TDF grid map} %
\label{TDFGridMap}

Notice that the previous kernel actually represents the $L_1$ TDF of an obstacle in its center. We can easily compute the TDF of a general point cloud as a grid map by just performing the bitwise-AND of this kernel with the grid map at each cell that contains a point. The representation of the distances in the form of a bit-set mask implicitly solves all the problems related to the distance field computation. Thus, we do not need to know if one point is close to any other, or the order of evaluation of points in the cloud. %Performing the bitwise-AND operation of the kernel centered in the point will automatically update the grid map considering such point, and also the existing information (distance to other existing points in the grid).

Another advantage of this approach is its easy parallelization. In principle, each point can update the grid map in parallel, and then merge all updates by means of bitwise-AND. We can make use of this feature to accelerate the computation of the TDF, distributing the computation among different threads in the CPU.

%Our TDF grid map will be represented as a 3D matrix of fixed resolution. The size of the grid is determined by the spatial distribution of 3D points. Every grid cell contains the $L_1$ distance to the closest point of the 3D map to that cell, or the truncation distance if the closest point is far.
\rev{We store the TDF map with two levels of hashing: the first level splits space into  $1\,\text{m}^3$  blocks, and the second level manages the high-resolution cells inside a block. Only blocks that actually receive points are expanded to high resolution (allocated on demand by the second level).} Each allocated fine cell stores the $L_1$ distance to the closest point of the 3D map to that cell, or the truncation distance if the closest point is far. The truncation distance will be defined by three factors: the number of bits used to represent the bit-set mask, the size of the kernel, and the grid resolution. The number of bits of the mask must be equal to the maximum $L_1$ distance into the kernel in order to make the approach efficient, so the truncation distance will be finally given by the size of the kernel and the resolution. %Let us consider a 3D kernel of $21 \times 21 \times 21$ as an example. We will make use of cubic kernels in this paper because the 3D world will be sampled evenly across all axes; however other structures are possible. In this example, the corners of the kernel are in coordinates $(10, 10, 10)$, $(10, 10, -10)$, $(10, -10, 10)$, etc. Thus, the maximum distance into this kernel is $L_1^{max}=30$ and, then, we need at least $30$ bits to encode the mask. Now, we can compute the truncation distance as the product of $L_1^{max}$ and the grid cell resolution. Assuming a cell resolution of $0.05m$, the truncation distance of this kernel is $1.5m$.

\subsection{Grid map initialization and access}

As previously introduced, \rev{we organize the map with a two-level (double) hashing scheme designed for memory efficiency. The first level partitions space into fixed $1\,\text{m}^3$ blocks. Many of these blocks will never receive measurements in typical scenes, so they remain lightweight: each block simply holds a pointer that is null until needed. When a point falls inside a block for the first time, that pointer is set to a newly allocated high-resolution container for that block. At the second level, the fractional part of the coordinates (scaled by the chosen resolution) determines the fine cell inside the block where the binary mask value is stored. This on-demand allocation ensures that only blocks receiving points are expanded to high resolution, while all other regions remain unallocated. If the maximum number of fine containers is reached, the oldest one is recycled through a circular buffer, keeping memory bounded and efficient.}

\rev{To ensure correct operation of the binary kernel with bitwise-AND, high-resolution containers are initialized to the maximum truncation distance, that is, a mask with all bits set. This value will be updated by the binary kernel as soon as the value to store into the grid cell is smaller than the truncation distance. Containers that remain unused are not allocated, and any recycled container (when hitting the memory cap) is re-initialized to the  maximum truncation distance before reassignment.}
%As we have a dense representation of the world, it is not recommended to use sparse representations such as kd-trees \cite{kdtree} or Octomap \cite{octomap}. Instead, we use a large memory array to represent the space at fixed resolution. A simple hash function is used to transform the 3D coordinates into the corresponding index in the array (see details in the source code). In this way, the time required for grid access is extremely small, and more importantly, constant. 

%Such grid map must be correctly initialized in order to properly work with the binary kernel bitwise-AND operation. Thus, the whole grid must be initialized to the truncation distance, that is, a mask with all bits set. This value will be updated by the binary kernel as soon as the value to store into the grid cell is smaller than the truncation distance.

\subsection{Distance computation}
\label{distancecomputation}
The grid map stores binary masks, not distances. We need a method to compute the actual $L_1$ distance from the binary mask in the grid cells. We make use of the C++ \emph{std::bitset()} standard function, which is directly mapped into computer assembly instructions. This function counts the number of bits set into a register, returning an integer with such number, the corresponding $L_1$ distance.
\subsection{Distance interpolation}
\label{trilinearinterpolation}
\rev{Following \cite{iros21dll}, we make use of trilinear interpolation in order to estimate the distance field beyond the grid cell resolution. %Trilinear interpolation is used as a trade-off between computation and accuracy. Thus, tricubic interpolation offers better accuracy, however its computation load in each grid cell prevents its online use.Using trilinear interpolation also offers an easy and quick distance field gradient approximation. Given such gradient approximation, we can analytically compute complex gradients with respect to third-party variables, as robot position or orientation for localization. 
However, here, the interpolation is computed online.}

%%%%%%%%%%%%%%%%%%%%%%%%%%%%%%%%%%%%%%%%%%%%%%%%%%%%%%%%%%%%%%%%%%%%%%%%%%%%%%%%
\section{Direct 3D LiDAR-Inertial Odometry}
\label{sec:dlo}

\begin{figure}[t!]
    \centering
    \includegraphics[width=0.97\linewidth]{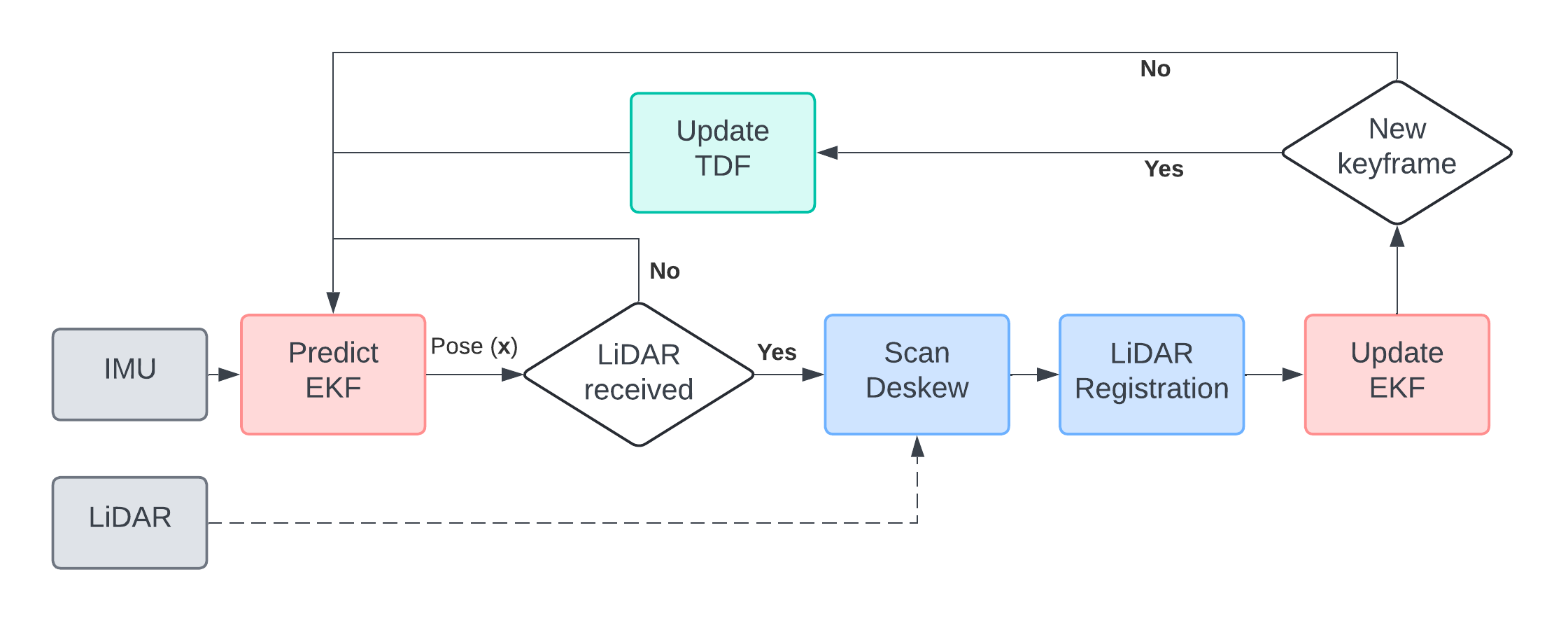}
    \caption{D-LIO Workflow. Red indicates the Kalman filter, green the TDF grid map, and blue the LiDAR preprocessing and optimization.}
    \label{dlo_scheme}
\end{figure}

Figure \ref{dlo_scheme} shows a block diagram of the D-LIO approach. The algorithm incrementally estimates the system pose using an Extended Kalman Filter (EKF) by integrating inertial measurements for pose prediction and point cloud to map registration for pose update. Thus, for each incoming point cloud, the system performs a cloud registration by aligning it with the environment mapped so far based on the Truncated Distance Field (TDF), using the EKF estimate as initial guess. This alignment is used to constrain the EKF's pose and velocity by means of state update. Whenever the estimated pose variation exceeds a predefined threshold, the TDF map is updated by incorporating new information. This updated map then facilitates a more precise alignment of subsequent point clouds during the registration process. %As a result, a sufficiently accurate initial pose estimate must be provided to ensure that the gradient-based optimization converges properly, leading to an improved pose estimate.

\subsection{Inertial State Integration}

Following LiDAR odometry common practices \cite{LIO-EKF}\cite{FastLIO2}, inertial sensors can be used to improve state estimation between scans. Inertial integration provides a good initial solution for LiDAR scan alignment, which benefits the subsequent scan-matching process. In addition, high-frequency inertial integration enables LiDAR scan deskewing, which also positively impacts the quality of the scan-matching process. We make use of an EKF with the following state vector:
\begin{equation}
    \mathbf{x} = [\mathbf{p},\mathbf{v}, \mathbf{a_b}, \mathbf{r}, \mathbf{g_b}]^t
\end{equation}
\noindent where the state encapsulates the 3D position $\mathbf{p}=(x, y, z)$, linear velocity \(\mathbf{v} = (v_x, v_y, v_z)\), accelerometer bias $\mathbf{a_b} = (\text{a}_{bx}, \text{a}_{by}, \text{a}_{bz})$, orientation $\mathbf{r}=(r_x, r_y, r_z)$ and gyroscope bias $\mathbf{g_b} = (\text{g}_{bx}, \text{g}_{by}, \text{g}_{bz})$. Here, the position $\mathbf{p}$, velocity $\mathbf{v}$, and orientation $\mathbf{r} $ are expressed in the global frame, while the accelerometer and gyroscope biases, $\mathbf{a_b}$ and $\mathbf{g_b}$  respectively, are expressed in the local inertial frame.

In the prediction step, the filter integrates IMU measurements (angular velocity and linear acceleration) to propagate the state forward in time. Once a new LiDAR scan is received, deskewing is performed using the high-rate inertial state estimates to compensate for motion-induced distortions accumulated during the scan acquisition. Subsequently, the update step is carried out using the refined pose from the cloud registration step and the estimated velocity derived from successive point cloud positions.
\subsection{LiDAR point-cloud registration}
This point cloud registration is formulated as a nonlinear optimization problem over the SE(3) space, where the objective is to minimize the distance from each point to the closest object in the map. The point-cloud is pre-transformed according to the EKF current state, so that the registration only needs to deal with the accumulated error in the inertial integration between scans. %Only those points whose initial transformation fall within the map boundaries are considered, ensuring that only valid observations contribute to the optimization.

The problem is formulated as the following minimization of a sum of residuals:
\begin{equation}
    \min_{\mathbf{t}, \mathbf{q}} \sum_{i}  \rho_i L_1(\mathbf{R}(\mathbf{q}) \cdot \mathbf{p}_i + \mathbf{t})^2
\end{equation}

\noindent where \textit{i} corresponds to each point in the cloud, $\mathbf{t} \in \mathbb{R}^3$ is the translation vector, $\mathbf{q} \in \mathbb{H}$ is the unit quaternion encoding rotation, $\mathbf{R}(\mathbf{q})$ is the corresponding rotation matrix, $\rho_i$ is the robust kernel detailed in Section \ref{sec:outlier}, and $L_1(\cdot)$  returns the unsigned $L_1$ distance to the nearest object using trilinear interpolation over the TDF computed so far as explained in Sections \ref{distancecomputation} and \ref{trilinearinterpolation}. 

The optimization is carried out using the Ceres Solver library \cite{Agarwal_Ceres_Solver_2022}. To properly handle the unit norm constraint of quaternions, a proper Lie group formulation is applied, along with its corresponding Lie algebra, enabling optimization in the tangent space.%of SO(3) while preserving the validity of the rotation representation.

\subsection{Outlier rejection}
\label{sec:outlier}
Before formulating the optimization problem, points that fall outside the TDF grid are discarded to ensure valid distance evaluations. 
For the remaining points, a robust loss function is applied to mitigate the effect of outliers while preserving global consistency in the estimation. In particular, a Cauchy loss is employed, whose scale parameter is a function of the point’s distance to the sensor:
\begin{equation}
    \rho_i = \text{CauchyLoss} \left( \lambda \cdot \left( 0.1 + 0.1 \cdot \| \mathbf{p}_i \| \right) \right)
\end{equation}
\noindent where $\lambda$ is a tuning factor and $\| \mathbf{p}_i \|$ denotes the Euclidean norm of the 3D point $\mathbf{p}_i$ expressed in the LiDAR sensor system reference. 

This robust kernel ensures that distant points are not excessively penalized due to their higher residuals, which can naturally arise from small orientation errors. For example, a rotation error of just $1^\circ$ may produce negligible displacements in nearby points (e.g., a few centimeters), but can lead to significant deviations (e.g., a few meters) in distant points. If a fixed-scale robust loss were used, these far points could be interpreted as outliers and ignored during optimization, causing the system to converge to a locally optimal but globally inaccurate solution. The proposed kernel preserves the contribution of both near and far points, promoting robust convergence across the entire cloud.

\subsection{Key-framing and map updating}
A keyframe-based approach is employed for map updates, where a new keyframe is created whenever the relative translation or rotation with respect to the previous keyframe exceeds the predefined thresholds $t_{\text{th}}$ and $q_{\text{th}}$, respectively.
When a new keyframe is created, the TDF map is updated by integrating the current point cloud, which is first transformed according to the latest optimized pose estimate, ensuring proper alignment in the global frame before fusion. This process, detailed in Section \ref{sec:ftedf}, updates the TDF grid by performing bitwise-AND operations of the kernel centered at each point, integrating it from the transformed cloud into the grid while considering both the new point and the existing map data. Although it is the most computationally expensive stage of the system, the use of bitwise-AND operations and parallelization helps optimize the map update process and significantly reduce processing time.

\rev{As noted earlier, in case all high-resolution containers are already in use when new points arrive (e.g., as the vehicle enters previously unseen areas), the system removes the information from the oldest $1\,\text{m}^3$ 
block and reassigns its high-resolution container to the newly observed block in the current scan, keeping memory bounded while prioritizing currently visible regions.}

%%%%%%%%%%%%%%%%%%%%%%%%%%%%%%%%%%%%%%%%%%%%%%%%%%%%%%%%%%%%%%%%%%%%%%%%%%%%%%%%

\section{\rev{Experimental results}} 
\label{sec:exp}

\rev{The algorithm has been tested on three datasets, the \textbf{VIRAL Dataset} \cite{nguyen2022ntu}, the \textbf{Newer College Dataset} \cite{zhang2021multicamera} \rev{and the \textbf{VBR Dataset} \cite{brizi2024vbrvisionbenchmarkrome}}, to evaluate its ability to correctly localize in different environments using diverse trajectories that challenge its performance.}

\rev{The accurate estimation of odometry has been primarily validated with the VIRAL Dataset, as it features more challenging environments and complex aerial trajectories that test the algorithm's stability. We evaluate the sequences ``eee",``nya", and ``sbs", using a DJI M600 Hexacopter equipped with two LiDARs (Ouster OS1-16 gen 1, horizontal and vertical), providing a combined density of approximately 20k points/scan. For the Newer College Dataset (``Quad-Easy" trajectory), we evaluate the reconstruction quality obtained from the TDF map, leveraging existing benchmarks on this dataset that report results across different approaches. Finally, the VBR dataset is used to demonstrate the algorithm’s scalability in a significantly larger environment, showcasing its ability to handle increasing map size and dense point clouds. In this case, we evaluate two trajectories: ``diag", an indoor/outdoor sequence of 1.4 km across three building floors with several loops, and ``pincio", an outdoor 2.5 km trajectory covering a wide open area with dense vegetation.  In both datasets, the platform is handheld with an Ouster OS-0-128 LiDAR; with 1024 columns for the Newer College (\(\sim\)130k points/scan) and 2048 columns for the VBR (\(\sim\)262k points/scan).}

%A supplementary video avaliable in the corresponding GitHub repository\footnote{https://anonymous.4open.science/r/D-LIO} presents qualitative results, 3D reconstructions, and the estimated trajectories.}

\rev{ Our results are presented in Tables \ref{VIRAL_ATE} and \ref{scalability} together with other approaches, such as A-LOAM \cite{9963800}, M-LOAM \cite{M-LOAM}, FAST-LIO2 \cite{FastLIO2}, KISS-ICP \cite{KISSICP}, LIO-SAM\cite{liosam2020shan} and PIN-SLAM \cite{Pan_2024}. \rev{In all three datasets, the Absolute Translation Error (ATE) was computed using the benchmarking tools provided by the VIRAL dataset to ensure a consistent and reliable trajectory evaluation. For the VBR dataset, we also report the Relative Pose Error (RPE), as it is especially informative for long trajectories}. The plots of all trajectories, together with complementary information and 3D reconstructions, are available in the supplementary video\footnote{https://www.youtube.com/watch?v=HmrA9YOCZ9w} and repository.\footnote{https://github.com/robotics-upo/D-LIO.git}} \rev{The experiments were performed on an HP Victus 16 laptop equipped with 32\,GB RAM and a 13th generation Intel Core i7-13700H processor. No GPU was used for the experimentation.}

\subsection{\rev{Trajectory evaluation}}
\label{evaluation}
\rev{These experimental results made use of Fast-TDF using 64 bit mask encoding together with a grid resolution of 0.05\,m, the truncation distance of this kernel is approximately 3 meters. The overall map extent is dataset-dependent; however, by using the proposed coarse layer of $1\,\text{m}^3$  blocks, we can scale the map footprint to very large areas at low resolution while allocating high-resolution cells only where points are observed. \rev{In our largest runs, we scaled the map to 500×500×100 m. % Hence, map extent is not the bottleneck; memory is dominated by the number of allocated high-resolution blocks, as detailed in Sec.~\ref{runtime}. 
Finally, the rotation/translation thresholds used to update the grid were set to \(\lambda=2.0\), \(t_{\text{th}}=1\,\mathrm{m}\), and \(q_{\text{th}}=\pi\,\mathrm{rad}\).}
As shown in Table~\ref{VIRAL_ATE}, our method attains the lowest ATE in most sequences and ranks second-best in the remainder, highlighting its consistent accuracy and robustness across diverse and challenging trajectories.} 

\rev{Please, notice that LIO-SAM \rev{and PIN-SLAM} performs back-end optimization, which makes the comparison with pure odometry not entirely fair. However, we have included its results to demonstrate that our approach, despite not using back-end optimization, achieves comparable performance.}

\begin{table}[t]
    \centering
    \caption{\rev{Experimental results on VIRAL and New College Dataset: \textit{ATE} (m) for all methods (m). The best result is highlighted in \textbf{bold} and the second best is \underline{underlined}, excluding LIO-SAM and PIN-SLAM as they perform back-end optimization. D-LIO, FAST-LIO2, and LIO-SAM are LiDAR–inertial methods, while the others rely solely on LiDAR. Metrics were taken from \cite{VIRAL-SLAM}, \cite{blancoclaraco2024flexibleframeworkaccuratelidar} and \cite{nvliom}.}}
    \resizebox{\columnwidth}{!}{ 
    \setlength{\tabcolsep}{2pt}
    \begin{tabular}{|c|>{\columncolor[gray]{0.95}}cc>{\columncolor[gray]{0.95}}cc>{\columncolor[gray]{0.95}}c|c>{\columncolor[gray]{0.95}}c|}
        \hline
        %\toprule
        %\cline{2-8}
        %\multicolumn{1}{@{}l|}{ }      & 
         & 
        \textbf{D-LIO}             & \textbf{A-LOAM}           & \textbf{M-LOAM} & \textbf{KISS-ICP}         & \textbf{FAST-LIO2}         & \textbf{LIO-SAM}  & \textbf{PIN-SLAM} \\ %& VIRAL \\
        % \midrule
        \hline
        \hline
        \textbf{eee\textsubscript{01}} & \textbf{0.074} & 0.212            & 0.249   & 2.383 &\underline{0.166}            & 0.075   & -\\ %& 0.066  \\
        \textbf{eee\textsubscript{02}} & \textbf{0.073} & 0.199            & 0.168 & 1.586 & \underline{0.100}             & 0.069   & 2.056\\ %& 0.063  \\
        \textbf{eee\textsubscript{03}} & \textbf{0.121} & 0.148            & 0.233  &   1.055 &\underline{0.142}            & 0.101   & 0.615\\ %& 0.037  \\
        \hline
        \textbf{nya\textsubscript{01}} & \underline{0.080} & \textbf{0.077}   & 0.123  & 0.359   & 0.127                      & 0.076   &0.086\\ %& 0.051  \\
        \textbf{nya\textsubscript{02}} & \underline{0.104}& \textbf{0.091}   & 0.191   & -  & 0.151                          & 0.07    & 0.092\\ %& 0.043  \\
        \textbf{nya\textsubscript{03}} & \textbf{0.065}   & \underline{0.108}& 0.226   & 1.389  & 0.130                      & 0.137   & 0.455\\ %& 0.032  \\
        \hline
        \textbf{sbs\textsubscript{01}} & \textbf{0.085}   &\underline{ 0.103} & 0.173  & 1.353  & 0.130                       & 0.089   & 0.205\\ %& 0.048  \\
        \textbf{sbs\textsubscript{02}} & \textbf{0.067}  &\underline{ 0.091} & 0.147   & 1.435 & 0.144                     & 0.083   & 0.528\\ %& 0.064  \\
        \textbf{sbs\textsubscript{03}} & \textbf{0.098}   & 0.367             & 0.153   & 1.037 & \underline{0.126 }         & 0.14    & 0.708\\ %& 0.054  \\
        \hline
        \textbf{q-easy}             & \underline{0.093}& \textbf{0.085}    & 0.141  & 0.100             & 0.100            & 0.074  &0.09 \\ %& -      \\
        % \bottomrule
        \hline
    \end{tabular}
    }
    \label{VIRAL_ATE}
\end{table}

\begin{figure*}[t!]
    \centering
    \begin{minipage}{0.2\textwidth}
        \centering
        \includegraphics[width=\textwidth]{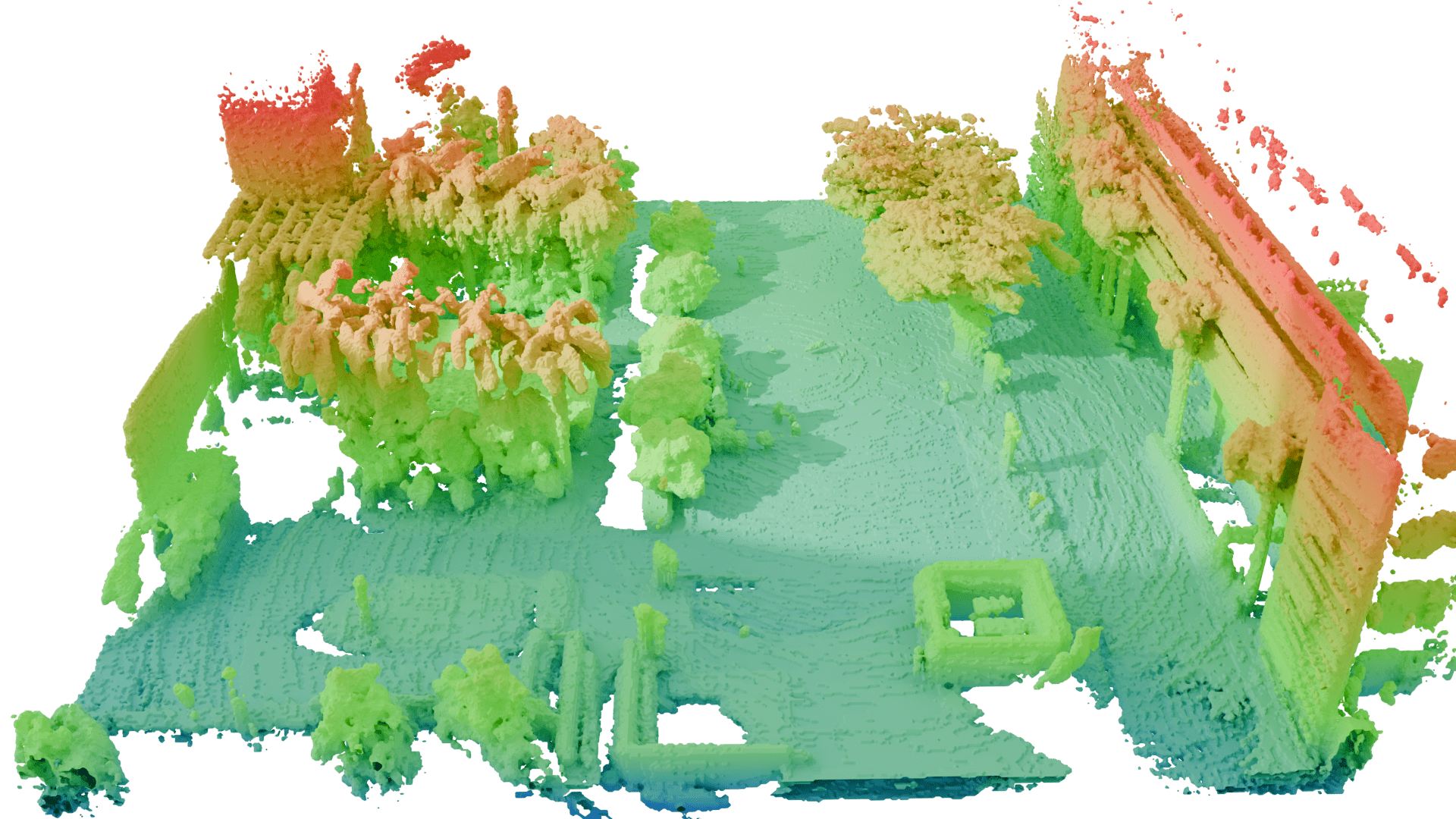}
    \end{minipage}%
    \hfill
    \begin{minipage}{0.2\textwidth}
        \centering
        \includegraphics[width=\textwidth]{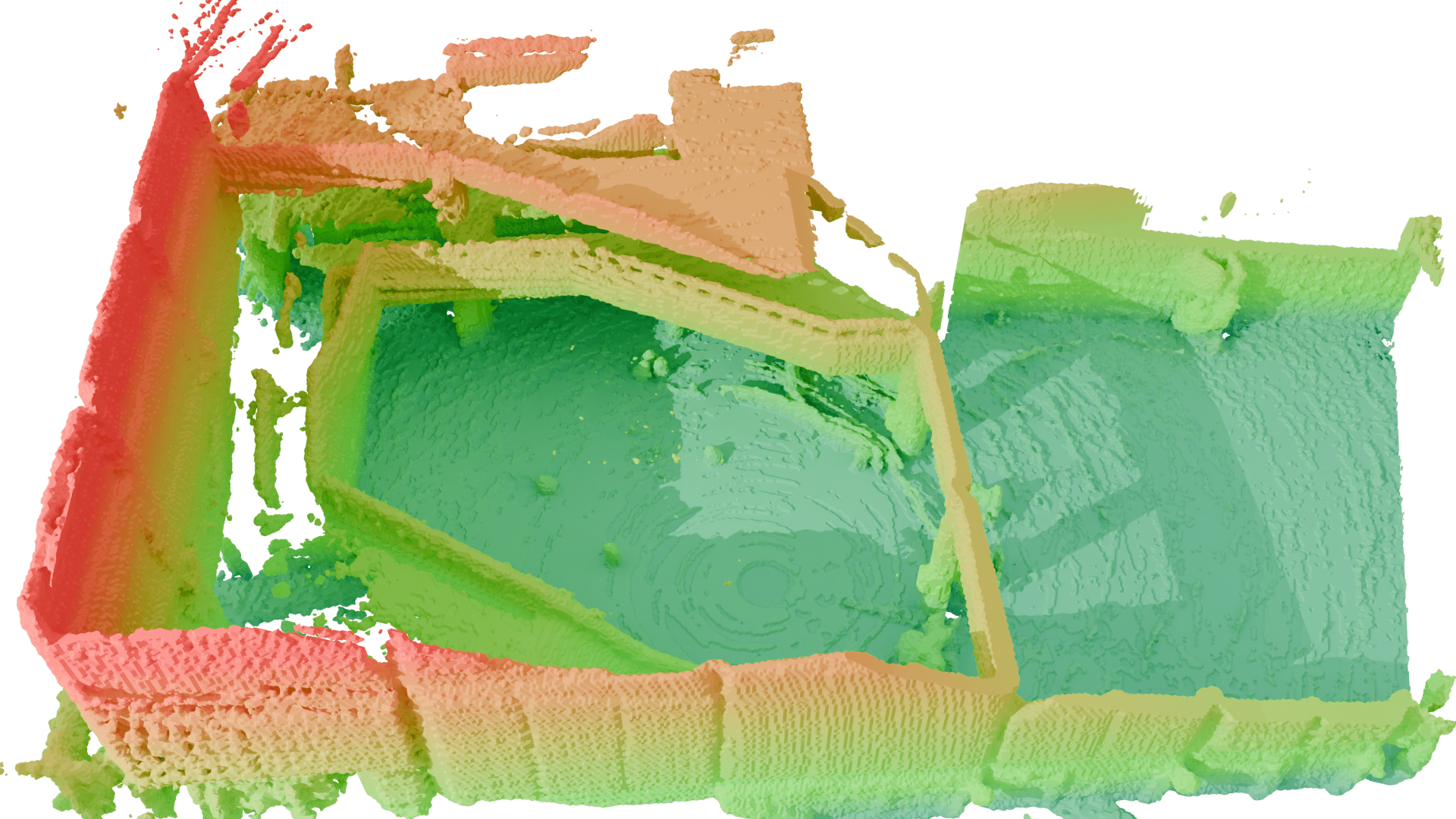}
    \end{minipage}%
    \hfill
    \begin{minipage}{0.2\textwidth}
        \centering
        \includegraphics[width=\textwidth]{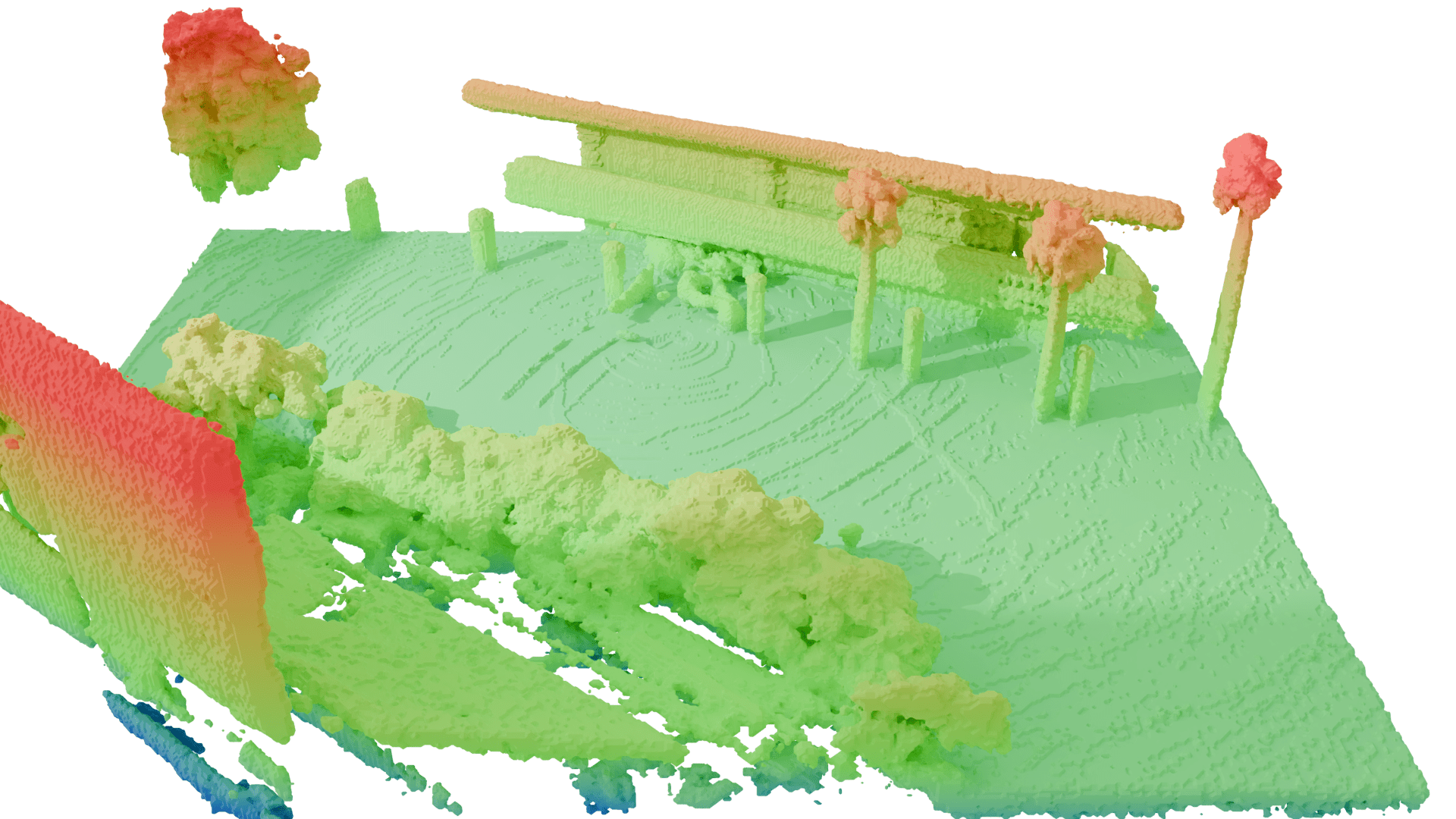}
    \end{minipage}%
    \hfill
    \begin{minipage}{0.2\textwidth}
        \centering
        \includegraphics[width=\textwidth]{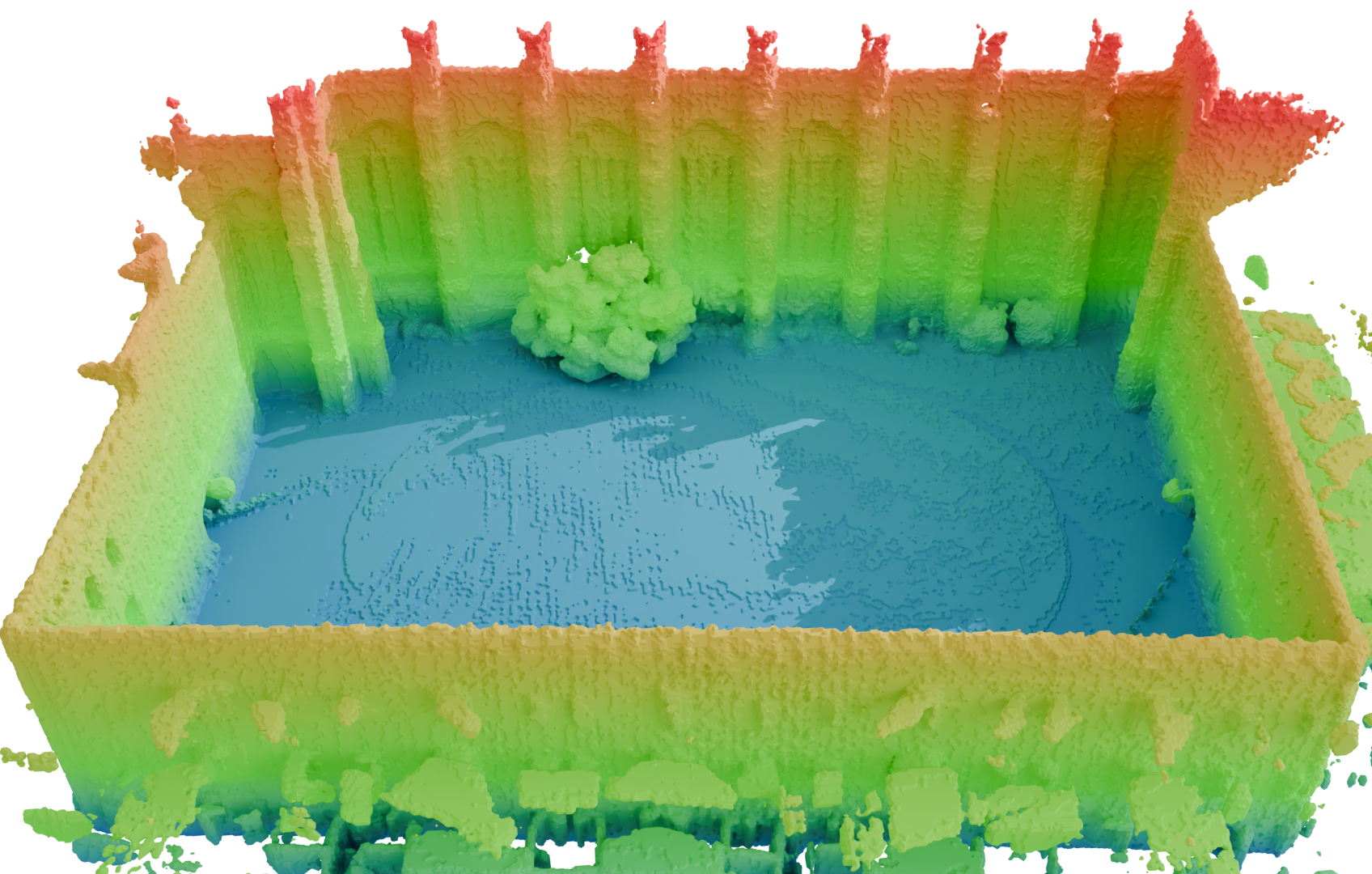}
    \end{minipage}
    \caption{3D map reconstructions (from left to right) of ``eee'', ``nya'', ``sbs'', and Newer College (Quad-Easy) datasets.}
    \label{fig:maps}
\end{figure*}

\subsection{\rev{Scalability}}

\rev{After validating the accuracy of our algorithm in more dynamic trajectories, we next assess scalability on longer trajectories and larger maps using the VBR Dataset (See Table \ref{scalability}). On the ``diag'' sequence, we allocated 250k high-resolution blocks within a 150$\times$150$\times$50\,m map. Despite the moderate map size, the trajectory is long (1.4\,km); however, the points are compactly distributed, so this allocation sufficed to represent the space without reallocations, as cells are created only around observed points, avoiding empty space and improving memory efficiency. Results are competitive with the state of the art, second only to PIN-SLAM, which benefits from global loop closures on long, loopy paths.}
\rev{In the ``pincio" sequence, the platform covers 2.5km over a much wider area. Since the full map cannot be kept in memory without reallocations, we limited the allocation to 50k high-resolution blocks. The system preserves detail only near the current pose, dynamically evicting distant blocks that no longer affect the current scan, thus managing memory dynamically. As this run spans a larger area and lacks loop closures, the absolute error increases; nevertheless, an ATE of 7\,m over 2.5\,km corresponds to only 0.28\% of the total trajectory length.}

\begin{table}[t]
\caption{\rev{Experimental results on VBR Dataset: ATE (m) and RPE (\%) in parenthesis for all methods. The best result is highlighted in \textbf{bold} and the second best is \underline{underlined}.}}
\label{scalability}
\centering

\begin{tabular}{|c|>{\columncolor[gray]{0.95}}cc>{\columncolor[gray]{0.95}}c|}
\hline
  & \textbf{D-LIO} & \textbf{KISS-ICP} & \textbf{PIN-SLAM} \\
\hline
\hline
\textbf{diag} & \underline{0.560} \scriptsize(0.72) & 1.397  \scriptsize (1.790) & \textbf{0.362} \scriptsize (0.468) \\
\textbf{pincio}  & 7.39 \scriptsize (3.45)  & \underline{0.784}  \scriptsize (0.485) &  \textbf{0.647} \scriptsize (0.453)  \\
\hline
\end{tabular}
\end{table}

\subsection{\rev{Mapping}}
\label{sec:mapping}

\rev{To quantitatively evaluate the mapping performance of our algorithm, we adopt standard metrics widely used for 3D reconstruction benchmarking, following the evaluation methodology proposed in \cite{Knapitsch2017}. The evaluation includes four key indicators: Chamfer-L1 distance (C-L1), reconstruction accuracy, completeness, and F-Score. Chamfer-L1 offers an aggregate measure integrating both reconstruction accuracy, representing the fidelity to the ground truth, and completeness, which reflects how thoroughly the environment has been captured. For accuracy, completeness, and Chamfer-L1 metrics, smaller values represent improved map quality. Conversely, higher F-Score values reflect a better balance between precision and recall. The detailed results of these metrics are summarized in Table \ref{tab:mapping-metrics}} 

\begin{table}[t!]
\caption{\rev{Mapping metrics on Newer College Quad-Easy sequence. Best and second best scores are marked in \textbf{bold} and \underline{underlined} respectively}}
\label{tab:mapping-metrics}
\begin{tabular}{|c|c|c|c|c|c|}
\hline
 & \textbf{Pose}                      & \textbf{Acc.}  & \textbf{Comp.} & \textbf{C-L1}  & \textbf{F-Score} \\ \hline\hline
\textbf{VDB-Fusion}   & \multirow{3}{*}{KissICP}  & 14.03 & 25.46 & 19.75 & 69.50   \\
\textbf{SHINE}        &                           & 14.87 & 20.02 & 17.45 & 68.85   \\
\textbf{NKSR}         &                           & 15.67 & 36.87 & 26.67 & 58.57   \\ \hline
\textbf{PUMA}         & \multirow{8}{*}{Odometry} & 15.30 & 71.91 & 43.60 & 57.27   \\
\textbf{SLAMesh}      &                           & 19.21 & 48.83 & 34.02 & 45.24   \\
\textbf{NeRF-LOAM}    &                           & 12.89 & 22.21 & 17.55 & 74.37   \\
\textbf{S²KAN-SLAM}   &                           & 13.32 & 18.80 & 16.06 & 72.03   \\
\textbf{ImMesh}       &                           & 15.05 & 19.80 & 17.42 & 66.87   \\
\textbf{PIN-SLAM}     &                           & \underline{11.55} & 15.25 & 13.40 & 82.08   \\
\textbf{KN-LIO}       &                           & \textbf{8.18}  & \underline{11.65} & \textbf{9.92}  & \textbf{91.01}   \\
\textbf{D-LIO} &                           & 12.14 & \textbf{10.63}  & \underline{11.39} & \underline{83.83}   \\ \hline
\end{tabular}
\end{table}

\rev{Overall, D-LIO remains at the state of the art: it achieves the best completeness (10.63), second-best Chamfer-L1 (11.39) and F-Score (83.83), and competitive accuracy (12.14). It approaches KN-LIO’s top performance and surpasses PIN-SLAM on Chamfer-L1, completeness, and F-Score, while classical TSDF/SDF and several odometry-driven baselines lag behind.}

\subsection{\rev{Runtime and memory consumption}}
\label{runtime}
\rev{The performance of the system was analyzed in terms of memory usage, CPU load, and execution time for all datasets. Overall memory consumption is governed by the number of high-resolution blocks (each a \(1\,\text{m}^3\) region discretized at 0.05\,m) rather than the coarse map, whose footprint is negligible by comparison (see Sec.~\ref{evaluation}). In VIRAL (largest trajectory), we used 180k blocks (\(31\%\) of RAM). In Newer College, due to the higher point concentration, only 45k blocks (\(22\%\) of RAM) were required. On \emph{diag}, despite a 1.4\,km path within one building, most points fell into already-activated regions; with an intentionally oversized budget of 250k blocks (\(47\%\) of RAM) we needed no reallocations. Finally, on \emph{pincio}, points were widely dispersed with little coarse-block reuse, so we ran a smaller allocation of 45k blocks (\(22\%\) of RAM) with dynamic reallocation as the platform advanced.}

\rev{Execution times are summarized in Table \ref{tab:map_runtime}. The map update step normally emerges as the most computationally expensive stage, as its time strongly depends on the point cloud density. For instance, in the VIRAL dataset, the lower density from OS-1 sensors ($\approx$ 20\,k points/scan) leads to shorter update times, whereas in the college and VBR dataset, the OS-0 sensor produces $\approx$ 100-260\,k points/scan making updates slower. However this cost also depends on environment topology: in diag, compact distributions and frequent revisits mean many already-seen regions are not updated, so the update time can match, or even be lower than, the optimization.}

\rev{Additionally, updates are triggered at a much lower frequency than optimization (which runs at the sensor rate, 10 Hz), so the weighted average tends to be driven by the optimization step, also influenced by the point density.  Note that all results were obtained under a worst-case computational load, no downsampling; in practice, applying voxel filtering can significantly speed up the system with negligible impact on accuracy.}

\rev{It is also worth noting that state-of-the-art distance field mapping approaches on CPU, such as VDB-GPDF \cite{VDB-GPDF}, which focuses solely on reconstruction, reported similar execution times to ours, but our approach performs both odometry and reconstruction without any downsampling. Although the map is updated less frequently, once every meter, it integrates the full-resolution point cloud, offering a balanced trade-off between computational efficiency and mapping accuracy.}
\begin{table}[t!]
\caption{\rev{Runtime analysis across different datasets with no downsample, showing mean ± standard deviation (in seconds) for each processing phase. Also reported, per trajectory, the average count of truly measured LiDAR returns.}}
\label{tab:map_runtime}
\centering
%\renewcommand{\arraystretch}{1.3}
%\setlength{\arrayrulewidth}{0.3mm}
%º\setlength{\tabcolsep}{10pt}
\begin{tabular}{|l|>{\columncolor[gray]{0.95}}c|c|>{\columncolor[gray]{0.95}}c|c|}
\hline
 & \textbf{Points} & \textbf{Total (s)} & \textbf{Optimize (s)} & \textbf{Update (s)} \\
\hline
\hline

\textbf{eee}& 20k    & 0.048 ± 0.055 & 0.046 ± 0.013 & 0.155 ± 0.009 \\
\textbf{nya} & 20k    & 0.049 ± 0.034 & 0.042 ± 0.014 & 0.146 ± 0.016 \\
\textbf{sbs} & 20k    & 0.051 ± 0.048 & 0.044 ± 0.016 & 0.142 ± 0.034 \\
\textbf{q-easy} & 100k & 0.479 ± 0.196 & 0.442 ± 0.158 & 0.470 ± 0.095 \\
\textbf{diag}  & 190k     & 0.603 ± 0.240 & 0.565 ± 0.194 & 0.315 ± 0.085\\
\textbf{pincio} & 140k     & 0.576 ± 0.256 & 0.505 ± 0.182 & 0.563 ± 0.146 \\

\hline
\end{tabular}
\end{table}
\section{CONCLUSIONS AND FUTURE WORK}
\label{sec:conclusions}
D-LIO, a fully direct LiDAR–inertial odometry pipeline based on Fast-TDF simultaneous mapping, has been shown to match or outperform feature-driven state-of-the-art methods while producing a continuously available distance field for downstream tasks. Unlike conventional \rev{Neural} TSDF/ESDF frameworks whose update cost grows with map volume, our Fast-TDF binary-mask scheme ties each integration step solely to the number of incoming points. By encoding obstacle proximity directly in the truncated distance field, we avoid feature extraction, yielding robust convergence even in feature-sparse or highly reflective environments. 

Future work will focus on %two main extensions. First, addressing the limitation of using a fixed-size grid, as there is currently no strategy to manage the map when the system approaches its spatial limits. Introducing dynamic resizing or shifting methods could improve scalability in larger environments. Second, 
extending the Fast-TDF kernel to support signed distance estimation, enabling the system to distinguish whether a point lies in front of or behind an obstacle, potentially enhancing registration accuracy.

%%%%%%%%%%%%%%%%%%%%%%%%%%%%%%%%%%%%%%%%%%%%%%%%%%%%%%%%%%%%%%%%%%%%%%%%%%%%%%%%
%\section*{Acknowledgment}
%\label{sec:Acknowledgment}
%This work is partially supported by the grants INSERTION (PID2021-127648OB-C31) and NORDIC (TED2021-132476B-I00), both funded by the “Agencia Estatal de Investigación – Ministerio de Ciencia, Innovación y Universidades” and the “European Union NextGenerationEU/PRTR”

\balance
\bibliographystyle{IEEEtran}

\end{document}